%% file: main.tex
\documentclass[runningheads]{llncs}
\usepackage{graphicx}
\usepackage{wrapfig}
\usepackage{array}
\graphicspath{ {./images/} }
\usepackage{amsmath,amssymb} 
\usepackage{color}
\usepackage{cite}
\usepackage{amsmath}
\usepackage{booktabs}
\usepackage[width=122mm,left=12mm,paperwidth=146mm,height=193mm,top=12mm,paperheight=217mm]{geometry}

\newcommand{\pbf}[1]{\vspace{-1.5mm}\paragraph{\textbf{#1 }}}
\begin{document}
\pagestyle{headings}
\mainmatter
\def\ECCVSubNumber{4218}  

\title{Semi-Local 3D Lane Detection and Uncertainty Estimation} 

\titlerunning{Semi-Local 3D Lane Detection and Uncertainty Estimation} 
\authorrunning{N. Efrat, M. Bluvstein, N. Garnett, D. Levi, S. Oron, B. Shlomo} 

\author{Netalee Efrat, Max Bluvstein, Noa Garnett, Dan Levi\\ Shaul Oron, Bat El Shlomo}
\institute{General Motors\\Advanced Technical Center Israel}

\maketitle

\begin{abstract}
\input{abstract}
\end{abstract}

\input{intro}
\input{related_work}
\input{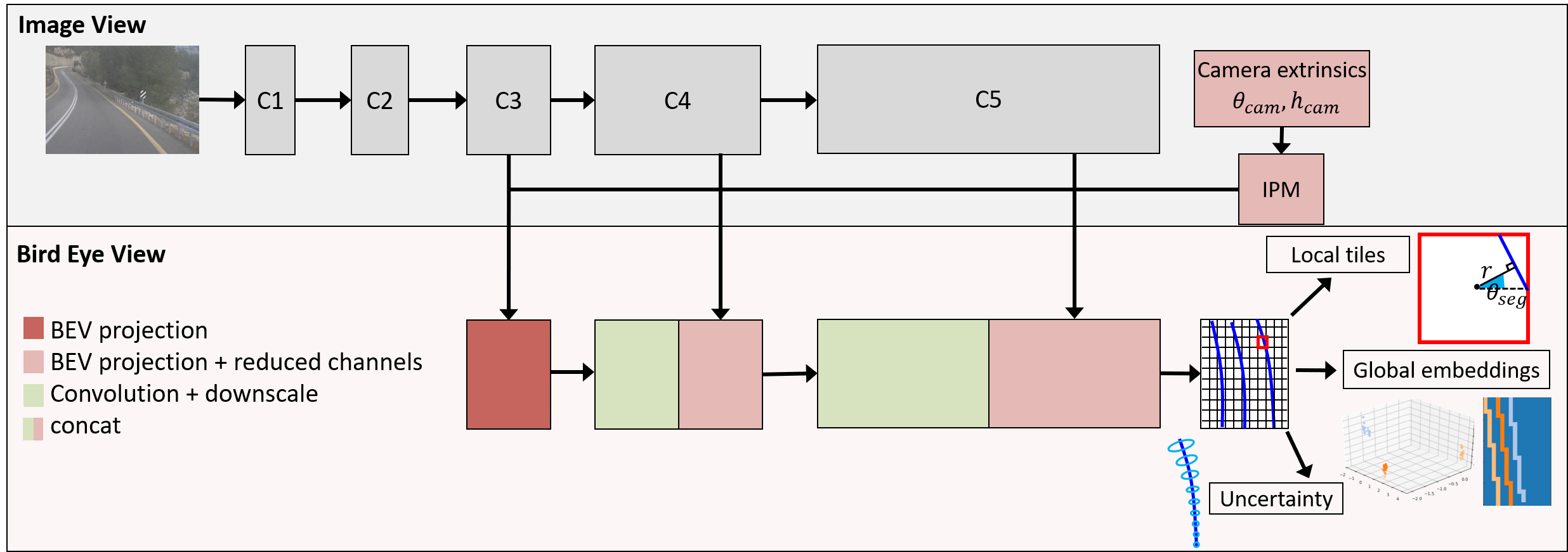}
\input{experiments}
\input{results}
\input{conclusions}

\clearpage

\bibliographystyle{splncs}
\bibliography{bibliography.bib}

\end{document}

%% file: abstract.tex
We propose a novel camera-based DNN method for 3D lane detection with uncertainty estimation. Our method is based on a semi-local, BEV, tile representation that breaks down lanes into simple lane segments. It combines learning a parametric model for the segments along with a deep feature embedding that is then used to cluster segment together into full lanes. This combination allows our method to generalize to complex lane topologies, curvatures and surface geometries. Additionally, our method is the first to output a learning based uncertainty estimation for the lane detection task. The efficacy of our method is demonstrated in extensive experiments achieving state-of-the-art results for camera-based 3D lane detection, while also showing our ability to generalize to complex topologies, curvatures and road geometries as well as to different cameras. We also demonstrate how our uncertainty estimation aligns with the empirical error statistics indicating that it is well calibrated and truly reflects the detection noise.  

%% file: intro.tex
\begin{figure}[hb]
\centering
\includegraphics[width=\textwidth]{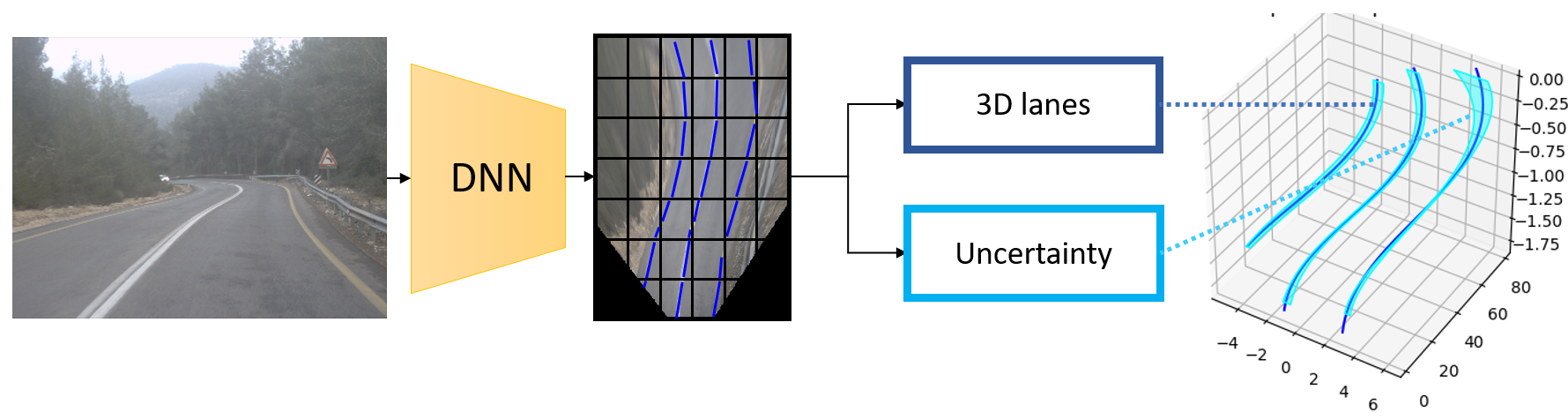}
\caption{Illustration of our camera-based 3D lane detection with uncertainty estimation network. Our method works in Bird Eye View perspective, which is rasterized to a coarse tiles grid. We output parametric 3D curve representations for all tiles, which are then processed to form entire 3D lane curves together with detection uncertainty estimates. See text for more details.}
\label{fig:figure1}
\vspace{-5mm}
\end{figure}

\section{Introduction}
\label{sec:intro}

Camera-based lane detection plays a central role in the semantic understanding of the world around a vehicle and can be used for many tasks including trajectory planning, lane keeping, vehicle localization and map generation. 

Driving related applications in general and specifically autonomous driving require 3D lane detection with uncertainty estimation that can generalize well to all kinds of lane topologies (e.g. splits, merges, etc.), curvatures, and complex road surface geometries. In addition, as autonomous driving is a safety-critical system, it depends upon reliable estimation of its detection noise, in our case, lane position uncertainty. This uncertainty allows downstream modules like localization and planning to be robust to errors by weighing in the uncertainty when using lane information. 

Despite the need for 3D lane detection, most existing methods \cite{TowardsEnd2End, pan2018SCNN, Hou_2019_ICCV, ELGAN, lee2017vpgnet} focus on 2D lane detection in the image plane. Additionally, existing methods typically support a limited number of lane topologies, mainly lane lines that are parallel to the vehicle direction of travel. Important topologies required for many driving scenarios such as splits, merges and intersection are for the most part not supported and disregarded. Another aspect not addressed in previous work is ensuring the lane detection system provides uncertainty estimates for its outputs. Recent work for object detection \cite{GussianYolo_2019_ICCV,monoloco, levi2020evaluating} suggest novel learning based methods for objects uncertainty estimation. However, to the best of our knowledge, none address a learning based solution for lane detection uncertainty.

In this work, we introduce a novel 3D lane representation and detection framework that is capable of detecting lanes, together with position uncertainty, for any arbitrary topology including splits, merges and lanes perpendicular to the vehicle travel direction. Our method generalizes to different road surface geometries and curvatures, as well as to different cameras. 

Key to our solution is a compact semi-local representation that is able to capture local topology-invariant lane structures and road surface geometries. Our lane detection is done in Bird's Eye View (BEV) which is divided into a regular grid of non-overlapping coarse tiles, as illustrated in Fig.  \ref{fig:figure1}. We assume lane segments passing through the tiles are simple and can be represented by a low dimensional parametric model.  Specifically, each tile holds a line segment parameterized by an offset from the tile center, an orientation and a height offset from the BEV plane. This semi-local tile representation lies on the continuum between global representation (entire lane) to a local one (pixel level). Each tile output is more informative than a single pixel in a segmentation based solution as it is able to reason on the local lane structure but it is not as constrained as the global solution which has to capture together the complexity of the entire lane topology, curvature and surface geometry.

Our representation breaks down lane curves into multiple lane segments but does not explicitly capture any relation between them. Adjacent tiles will have overlapping receptive fields, and thus correlated results, but the fact that several tiles represent the same lane entity is not captured. In order to generate full lane curves we learn an embedding for each tile which is globally consistent across the lane. This enables clustering small lane segments into full curves. As we show in our experiments (Sec. \ref{sec:results}), the combination between the semi-local tile representation and the embedding based clustering allows the network to output full 3D lane curves of any topology or surface geometry.

Another key component of our method is its ability to provide a noise estimate for the detected lane positions. This uncertainty estimation is achieved by modeling the network outputs as Gaussian distributions and estimating their mean and variance values. This is done for each lane segment parameter and then combined together to produce the final \textit{Covariance} matrix for each lane point.
Unlike the segment parameters that can be learned locally across tiles, the empirical errors required for training the uncertainty depend on all the tiles composing an entire lane and have to be reasoned globally as will be further explained in Sec. \ref{sec:method_uncertainty} and shown in our experiments. To the best of our knowledge, this is the first learning based uncertainty estimation method for lane detection.

We run extensive experiments, using three datasets, that show our method improves the average precision ($AP$) over the current 3D camera-based state-of-the-art 3D-LaneNet \cite{garnett20183dlanenet} by large margins. We demonstrate qualitatively and quantitatively the efficacy of our learning based clustering, and our method generalization to new lane curvatures and surface geometries as well as new cameras and unseen data. Finally, we present our learning based lane position uncertainty results, and show that it can properly capture the statistics of the actual error of our predicted lanes.

To summarize, the main contribution of our work is twofold: (a) We present a novel 3D semi-local lane representation and detection framework that generalizes to arbitrary lane topologies, curvatures and road surface geometries as well as different camera setups. (b) We propose the first learning based method to provide position uncertainty estimation for the lane detection task.

%% file: related_work.tex
\section{Related work}\label{sec:related_work}

\pbf{2D lane detection} Most existing lane detection methods focus on lane detection in the image plane and are mostly limited to parallel lane topologies. The literature is vast and includes methods performing 2D lane detection by using self attention \cite{Hou_2019_ICCV}, employing GANs \cite{ELGAN}, using new convolution layers \cite{pan2018SCNN}, exploit vanishing points to guide the training \cite{lee2017vpgnet} or use differentiable least-squares fitting \cite{wvangansbeke_2019}. Most related to ours is the method of \cite{DeepLearningHW_AndrewNg} that uses a grid based representation in the image plane, with a line parametrizations and density based spatial clustering for highway lane detection. Our approach uses BEV and a different parametrization than \cite{DeepLearningHW_AndrewNg} and performs 3D lane detection. We also use learning based clustering as well as output uncertainty estimates for detected lanes.
Another work related to ours is \cite{TowardsEnd2End} that uses learned embedding to perform lane clustering. While \cite{TowardsEnd2End} perform segmentation at the image pixel level, we cluster the lane segments in BEV on the semi-local tile scale, which is far less computationally expensive.

\pbf{3D lane detection} Detecting lanes in 3D is a challenging task drawing increasing attention in recent years. 3D lane detection methods can be roughly divided to LiDAR-based methods, camera-based and hybrid-methods using both like in \cite{bai2018deep}. In that work a CNN uses LiDAR to estimate road surface height and then re-projects the camera to BEV accordingly. The network doesn't detect lane instances end-to-end, but rather outputs a dense detection probability map that needs to be further processed and clustered. 
More related to ours are camera-based methods. The DeepLanes method \cite{gurghian2016deeplanes} uses a BEV representation but works with top-viewing cameras that only detect lanes in the immediate surrounding of the vehicle without providing height information. Another work closely related to ours is 3D-LaneNet \cite{garnett20183dlanenet} which also performs camera-based 3D lane detection using BEV. However, unlike our semi-local representation they use a global description that relies on strong assumptions regarding lane geometry, and therefore are unable to detect lanes which are not roughly parallel to ego vehicle direction, lanes starting further ahead, and other non-trivial topologies as will be shown in our experiments (Sec. \ref{sec:results}). 

\pbf{Uncertainty estimation} Despite its importance, uncertainty estimation for lane detection is not addressed in the literature. We therefore review work done on uncertainty estimation for object detection and classification. To estimate the prediction uncertainty during inference, the machine learning module should output a full distribution over the target domain. Among the available approaches are Bayesian neural networks \cite{GalThesis16,Gal_Ghahramani16}, ensembles \cite{Lakshminarayanan17} and outputting a parametric distribution \textit{directly} \cite{Nix94, levi2020evaluating}. In addition, since uncertainty estimates rely on observed errors on the \textit{training set}, they are often underestimated on the test set and require post training re-calibration. In the context of on-road perception, several works estimate uncertainty in object localization~\cite{levi2020evaluating, Phan18} but as far as we know, no previous work applies such techniques to lane detection. In this work we follow~\cite{levi2020evaluating} which provides a practical solution to both uncertainty estimation and re-calibration. Moreover, as further discusses in Sec. \ref{sec:method_uncertainty}, uncertainty estimation for general curves requires additional reasoning to that of object localization, such that properly reflects the error of each locally estimated segment with its associated global lane entity. 

%% file: method.tex
\section{Lane detection and uncertainty estimation}\label{sec:method}
\begin{figure}[!htb]
\vspace{-2mm}
\centering
\includegraphics[width=\textwidth]{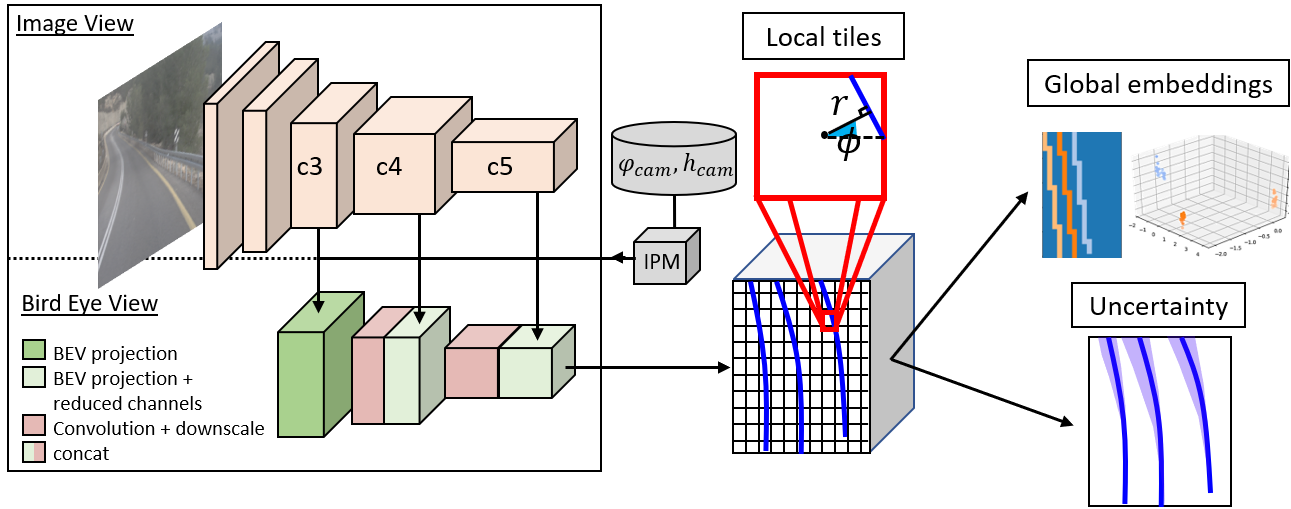}
\caption{Method overview. Our network is comprised of two processing pipelines: in image view (top) and in Bird Eye View (bottom). The image view encoder is composed of resnet blocks each one multiplying the number of channels. The BEV backbone is comprised of projected image view feature maps which are concatenated with the convoluted projected feature map from the former block. The final decimated BEV feature map is the input to the lane prediction head which outputs local lane segments, global embedding for clustering the segments to entire lanes, and lane point position uncertainty which relies both on the local tiles and on the entire lane curves.}
\label{fig:method}
\vspace{-5mm}
\end{figure}

We now describe our 3D lane detection and uncertainty estimation framework. A schematic overview appears in Fig.  \ref{fig:method}. We first present our semi-local tile representation and lane segment parameterization (Sec. \ref{sec:method_representation}) followed by how lane segments are clustered together using a learned embedding (Sec. \ref{sec:method_embedding}). Next, we discuss how uncertainty is estimated and calibrated (Sec. \ref{sec:method_uncertainty}) and finally how the lane structure is inferred from the network's output (Sec. \ref{sec:method_output}). 

\subsection{Learning 3D lane segments with Semi-local tile representation}\label{sec:method_representation}
Lane curves have many different global topologies and lie on road surfaces with complex geometries. This makes reasoning for entire 3D lane curves a very challenging task. Our key observation is that despite this global complexity, on a local level, lane segments can be represented by low dimensional parametric models. Taking advantage of this observation we propose a semi-local representation that allows our network to learn local lane segments thus generalizes well to unseen lane topologies, curvatures and surface geometries. 

\begin{wrapfigure}{r}{0.5\textwidth}
\includegraphics[width=0.48\textwidth]{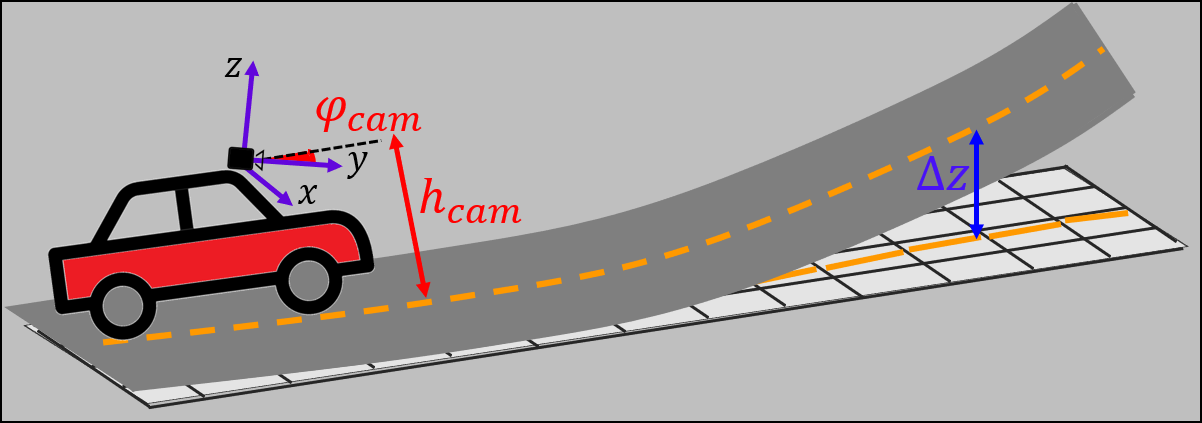}
\caption{The road projection plane is defined according to the camera mounting pitch angle $\varphi_{cam}$ and height $h_{cam}$, hence our representation is invariant to the camera extrinsics. We represent the GT lanes in full 3D relatively to that plane.}
\label{fig:projection_plane}
\vspace{-3mm}
\end{wrapfigure}

The input to our network is a single camera image. We adopted the dual pathway backbone proposed by Garnett et al. \cite{garnett20183dlanenet} which uses an encoder and an Inverse Perspective Mapping (IPM) module to project feature maps to Bird  Eye  View (BEV). The projection applies a homography, defined by camera pitch angle $\varphi_{cam}$ and height $h_{cam}$, that maps the image plane to the road plane (see Fig.  \ref{fig:projection_plane}). The final decimated BEV feature map is spatially divided into a grid $G_{W\times H}$ comprised of $W\times H$ non-overlapping tiles. Similar to \cite{garnett20183dlanenet}, the projection ensures each pixel in the BEV feature map corresponds to a predefined position on the road, independent of camera intrinsics and pose. 

We assume that through each tile $g_{ij}\in G_{W\times H}$ can pass a single line segment which can be approximated by a straight line. Specifically, the network regresses, per each tile $g_{ij}$, three parameters: lateral offset distance relative to tile center $\widetilde{r}_{ij}$, line angle $\widetilde{\phi}_{ij}$, (see Local tiles in Fig.  \ref{fig:method}) and height offset $\widetilde{\Delta z}_{ij}$ (see Fig.  \ref{fig:projection_plane}). In addition to these line parameters, the network also predicts a binary classification score $\widetilde{c}_{ij}$ indicating the probability that a lane intersects a particular tile. GT regression targets for the offsets and angles are calculated by approximating the lane segments intersecting the tiles to straight lines using the GT lane points after they were projected to the road plane (Fig. \ref{fig:projection_plane}).    

Position and $z$ offsets are trained using an $L1$ loss: 
\begin{equation}
\mathcal{L}_{ij}^{Offsets} = \|\widetilde{r}_{ij} - r_{ij}\|_1 + \|\widetilde{\Delta z}_{ij} - \Delta z_{ij}\|_1
\end{equation}

Predicting the line angle $\widetilde{\phi}_{ij}$ is done using the hybrid classification-regression framework of \cite{Mahendran2018AMC} in which we classify the angle $\phi$ (omitting tile indexing for brevity) to be in one of $N_{\alpha}$ bins, centered at $\alpha = \{ \frac{2\pi}{N_{\alpha}}\cdot i \}_{i=1}^{N_{\alpha}}$. In addition, we regress a vector $\Delta^{\alpha}$, corresponding to the residual offset relative to each bin center. Our angle bin estimation is optimized using a soft multi-label objective, and the GT probabilities are calculated as $p^{\alpha} = [{1 - |\frac{2\pi}{N_{\alpha}}\cdot i - \phi| / \frac{2\pi}{N_{\alpha}}}]_+ $. 
The GT offsets $\Delta^{\alpha}$ are the difference between the GT angle and the bin centers, and their training is supervised on the GT angle bin and adjacent bins to ensure that the delta offset can account for erroneous bin class prediction. 

The angle loss is the sum of the classification and offset regression losses:
\begin{multline}
    \mathcal{L}^{angle}_{ij} = \sum_{\alpha=1}^{N_{\alpha}}
    {[p^{\alpha}_{ij}\cdot \log{\widetilde{p}^{\alpha}_{ij}} + (1- p^{\alpha}_{ij}) \cdot \log{(1 - \widetilde{p}^{\alpha}_{ij})}  + \delta^{\alpha}_{ij} \cdot \|\widetilde{\Delta}^{\alpha}_{ij} - \Delta^{\alpha}_{ij} \|_1]}
\end{multline}
where $\delta^{\alpha}_{ij}$ is the indicator function masking the relevant bins for the offset learning.

The lane tile probability $\widetilde{c}_{ij}$ is trained using a binary cross entropy loss:
\begin{equation}
    \mathcal{L}^{score}_{ij} = c_{ij} \cdot\log{\widetilde{c}_{ij}} + (1-c_{ij}) \cdot\log{(1 - \widetilde{c}_{ij})}
\end{equation}

Finally, the overall tile loss is the sum over all the tiles in the BEV grid:
\begin{equation}
    \mathcal{L}^{tiles} =  \sum_{i,j\in W\times H}( 
    \mathcal{L}^{score}_{ij} +c_{ij} \cdot \mathcal{L}^{angle}_{ij} +  c_{ij}\cdot \mathcal{L}^{offsets}_{ij})
    \label{Eq:loss_tiles}
\end{equation}

\subsection{Global embedding for lane curve clustering}\label{sec:method_embedding}
In order to provide complete lane curves we need to cluster together multiple lane segments into complete lane entities.  To this end, we learn an embedding vector $f_{ij}$ for each tile such that vectors representing tiles belonging to the same lane would reside close in embedded space while vectors representing tiles of different lanes would reside far apart. For this we adopted the approach of \cite{TowardsEnd2End, Semantic_Instance_Segmentation}, and use a discriminative push-pull loss. Unlike previous work, we use the discriminative loss on the decimated tiles grid, which requires far less computations than operating at the pixel level. 

The discriminative push-pull loss is a combination of two losses:
\begin{equation}
    \mathcal{L}^{embedding} = \mathcal{L}^{pull} + \mathcal{L}^{push}
\end{equation}
A pull loss aimed at pulling the embeddings of the same lane tiles closer together:
\begin{equation}
    \mathcal{L}^{pull} = \frac{1}{C}\sum_{c=1}^C\frac{1}{N_c}\sum_{ij\in W\times H}{
    [\delta^c_{ij}\cdot\|\mu_c - f_{ij}\| - \Delta_{pull}]_+^2
    }
    \label{Eq:pull}
\end{equation}
and a push loss aimed at pushing the embedding of tiles belonging to different lanes farther apart: 
\begin{equation}
    \mathcal{L}^{push} = \frac{1}{C(C-1)}\sum_{c_{A}=1}^C\sum_{c_{B}=1,c_{B}\neq c_{A}}^C{ [\Delta_{push} - \|\mu_{c_{A}} - \mu_{c_{B}}\|]^2_+ }
    \label{Eq:push}
\end{equation}

where $C$ is the number of lanes (can vary), $N_c$ is the number of tiles belonging to lane $c$, $\delta_{ij}^c$ indicates if tile $i,j$ belongs to lane $c$, $\mu_c=\frac{1}{N_c}\sum_{ij\in W\times H}{\delta^c_{ij}\cdot f_{ij} }$ is the average of $f_{ij}$ belonging to lane $c$, $\Delta_{pull}$ constraints the maximal inter-cluster distance and $\Delta_{push}$ is the intra-cluster minimal required distance. 

Given the learnt feature embedding we can use a simple clustering algorithm to extract the tiles that belong to individual lanes. We adopted the clustering methodology from Neven et al. \cite{TowardsEnd2End} which uses mean-shift to find the clusters centers and set a threshold around each center to get the cluster members. We set the threshold to $\frac{\Delta_{push}}{2}$.

\subsection{Uncertainty estimation} \label{sec:method_uncertainty}
We now explain how we estimate the noise (uncertainty) of our lane detector. As it is a statistical property, its estimation is done by casting the tile prediction problem as a distribution estimation task. This means that we formulate each one of the lane segment parameters (omitting tile indexing for brevity) $y \in \{r, \phi, \Delta z\}$  as  a Gaussian distribution such that
\begin{equation}
    y|x,\theta \sim \mathcal{N}(\mu(x,\theta), \sigma^2(x,\theta))
\end{equation}

where $x$ is the network input and $\theta$ the network parameters. The mean values of the above distributions are the predicted values for the tile parameters i.e. $\mu_{r}(x,\theta) = \widetilde{r},\: \mu_{\phi}(x,\theta) = \widetilde{\phi},\: \mu_{\Delta z}(x,\theta) = \widetilde{\Delta z}$, estimated using the methodology described in Sec. \ref{sec:method_representation}. In this section we focus on the estimation of the variances, given the predicted mean values, as a second training stage. The variances $\sigma^2(x,\theta)$ are estimated through the optimization of the Negative Log Likelihood (NLL) which is a standard measure for probabilistic models quality \cite{direct_parametric_uncertainty, NIPS2017_7219}

\begin{equation}\label{eq:uncertainty}
NLL = -\log{p(y|x,\theta)} = \frac{1}{2} \log{\sigma^2(x,\theta)} + \frac{(y-\mu(x,\theta))^2}{2\sigma^2(x,\theta)} + const.
\end{equation}

\begin{wrapfigure}{r}{0.5\textwidth}
\centering
\includegraphics[width=0.48\textwidth]{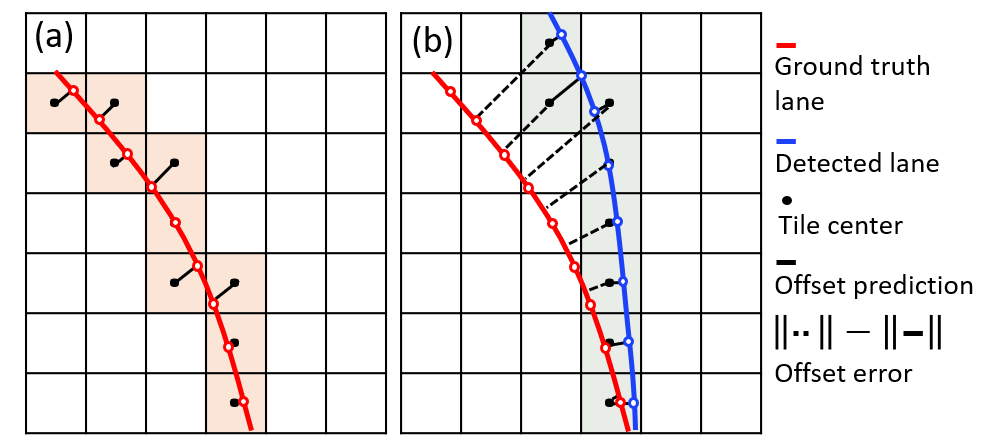}
\caption{Error calculation for uncertainty prediction supervision. (a) The orange tiles are occupied by the ground truth segments. On these tiles we supervise the tiles parameters prediction, and they can be easily extended to output those parameters uncertainty based on the within tile error. (b) By calculating the error while considering the actual network predictions (green tiles), together with the entire curve, we can better quantify the detector uncertainty.}
\label{fig:tiles_MSE}
\vspace{-3mm}
\end{wrapfigure}

where $y$ is the GT value and $(y-\mu(x,\theta))^2$ is the empirical Squared Error (SE). Proper evaluation of the empirical SE is therefore a key component in quantifying the uncertainty. Measuring the SE for the predicted tiles is not trivial because it is not obvious which is the corresponding GT segment the error should be calculated with respect to. This correspondence depends not only on the predicted tile itself, but on the entire curve it belongs to. Therefore, we propose measuring the SE in a global context of full lane curves as illustrated in Fig.  \ref{fig:tiles_MSE}b. Alternative solution for measuring the SE would be to simply calculate the error on the same tiles supervised for the tile parameters prediction, i.e., the tiles for which $c_{ij}=1$ (Fig.  \ref{fig:tiles_MSE}a). Using this solution, the errors originate only from the semi-local tile context and essentially bounded to the tile size thus would generate a skewed statistic. 
Obtaining the global-context error values requires that we first cluster lane segments together into full lane curves and then associate them to GT lanes. Once this is performed we can find, for each predicted lane segment, a corresponding GT segment \textit{in the context of the full associated GT lane} that can now be far from its semi-local context. This makes the uncertainty training a multi-stage process that requires first inferring lane segment parameters on the tile level $\mu(x,\theta)$, then clustering tiles together into full lanes, associating these lanes to GT lanes, computing the SE of the lane segments with respect to the associated GT segments, and finally using these SE values to compute the NLL loss and update the network parameters. 

Despite the supervision involved in the training process, deep neural networks tend to produce over confident uncertainty predictions (high probabilities, and small variances) \cite{pmlr-v70-guo17a_OnCalibration, levi2020evaluating}. It is therefore useful to further calibrate\footnote{Calibrated uncertainty is broadly defined as having the variance reflect the actual MSE error of the predictions in case of regression, and the probability output to match the actual accuracy in case of classification.} the uncertainty estimation of the network after it is trained. For this we adopt the Temperature Scaling \cite{pmlr-v70-guo17a_OnCalibration, levi2020evaluating} procedure and use it to calibrate the tiles classification scores and regressed offsets and angles uncertainties. Temperature scaling uses a single scalar parameter $T$ per parameter ($r, \phi, \Delta z$) that multiplies the estimated variance such that  $\sigma_T^2(x,\theta) = T \cdot \sigma^2(x,\theta)$. During the calibration training the $NLL$ is optimized with $\sigma_T^2(x,\theta)$ but only $T$ is updated. The training is performed on a different train set and the parameter $T$ adjusts the learnt variances to better capture the statistics on the new dataset. 

\subsection{Final output}\label{sec:method_output}
Our 3D lane detection module outputs lanes represented as sets of 3D directed lane points, where each lane point has an associated 3D covariance matrix indicating its estimated uncertainty, and local direction vector based on $\widetilde{\phi}_{ij}$.  

Lane segments are clustered together as explained in Sec. \ref{sec:method_embedding}. Each lane segment (tile) contributes a point to the lane point set. We begin by thresholding tile score to output only tiles with lanes. We then 
convert the offsets ($\widetilde{r}_{ij}, \widetilde{\Delta z}_{ij}$) and angles ($\widetilde{\phi}_{ij}$) to points by converting them from Polar to Cartesian coordinates. Finally, we transform the points from the BEV plane to the camera coordinate frame by subtracting $h_{cam}$ and rotating by $-\varphi_{cam}$ (see Fig.  \ref{fig:projection_plane}).
\begin{equation}
\begin{bmatrix}
\widetilde{x}_{ij} \\
\widetilde{y}_{ij} \\
\widetilde{z}_{ij}
\end{bmatrix} = 
\begin{bmatrix}
1 & 0 & 0\\
0 & \cos(\varphi_{cam})  & \sin(\varphi_{cam}) \\
0 & -\sin(\varphi_{cam}) & \cos(\varphi_{cam})
\end{bmatrix} \cdot
\begin{bmatrix}
\widetilde{r}_{ij} \cdot \cos(\widetilde{\phi}_{ij}) \\
\widetilde{r}_{ij} \cdot \sin(\widetilde{\phi}_{ij}) \\
\widetilde{\Delta z}_{ij} - h_{cam}
\end{bmatrix}
\label{Eq:interpret}
\end{equation}

Although the network learns to predict the variances of offsets and angles independently, we output a full covariance matrix for each predicted lane point in Cartesian coordiantes.
This is done by transforming the offsets and angle covariance matrix $Cov^{pol}_{ij}=diag[\widetilde{\sigma}_{r}^2, \widetilde{\sigma}_{\phi}^2, \widetilde{\sigma}_{\Delta z}^2]_{ij}$ to Cartesian space:
\begin{equation}
    Cov^{cart}_{ij} = J_{ij} \cdot Cov^{pol}_{ij} \cdot J_{ij}^T
\end{equation}
Where $J_{ij}$ is the Jacobian matrix of the Polar to Cartesian conversion in Eq. \eqref{Eq:interpret} approximated for segment $i,j$.

%% file: experiments.tex
\section{Experimental Setup}\label{Sec:Experiments}
We study the performance of our 3D lane detection framework using several 3D-lane datasets, comparing it to \cite{garnett20183dlanenet} which is the current state-of-the-art (SOTA) camera-based 3D-lane detector. We demonstrate the method ability to detect difficult lane topologies and generalize to complex surface geometries and different cameras setups. Finally, we show the accuracy of our uncertainty estimation.  

\pbf{Datasets } Evaluation is done using two 3D-lane datasets. The first is \textit{synthetic-3D-lanes} \cite{garnett20183dlanenet} containing synthetic images of complex road geometries with 3D ground truth lane annotations. The second, is a dataset  we collected and annotated\footnote{Annotation protocol is similar to that used in \cite{garnett20183dlanenet}.} referred to as \textit{3D-lanes}. 
This dataset contains $327K$ images from 19 distinct recordings (different geographical locations at different times) taken at 20 fps. The data is split such that train and test sets have different distributions. Specifically, the train set, $298K$ images, is comprised mostly of highway scenarios while the test set is comprised of a rural scenario with complex curvatures and road surface geometries, taken at a geographic location not in the train set. 
To reduce temporal correlation we sampled every 30'th frames giving us a test set of 1000 images. We also set aside $5K$ images for uncertainty calibration training. Example images from the train and test sets can be seen in Fig. \ref{fig:real_data}.

To quantitatively demonstrate our methods ability to generalize to new cameras and scenes we use the tuSimple 2D benchmark \cite{tusimple_dataset}. Additional qualitative evaluation using a new camera setup is also shown.
  
\pbf{Evaluation} We adopt the AP metric commonly used in object detection \cite{ms_coco, Geiger2012CVPR} that averages the area under ROC curves, generated with different IOU thresholds. Unlike object detection, where intersection and union for bounding boxes are well defined, intersection and union for 3D curves are not. Similarly to \cite{DAGMapper_2019_ICCV} we define the intersection for curves as the length of the curve sections that are closer than a threshold to the GT curve, and the union as the length of the longer curve out of the two: detected and GT curves. However, unlike [26], we calculate the True Positive (TP), False Positive (FP) and miss detections, for every lane curve, not for every lane point. This gives a better estimate of the number of lanes properly detected, regardless of their length, distance, or topology (merges, splits or intersections). For example, in per-point metrics, detecting half of the points out of two lanes, or detecting only one lane out of the two, would get the same score, whereas per-lane metrics will give these two cases different scores.
Note that in order to determine if a certain lane section intersects or not, we
have to define a distance threshold. This is a heuristic that does not exist when defining IOU in object detection. To this end we add another set of distance accuracy metrics to account for the location error of each detected lane point with respect to its associated GT curve. We divided the entire dataset to lane points in the near range (0-30m) and far range (30-80m) and calculate the mean absolute lateral error for every range.

\pbf{Implementation details } We use the dual-pathway architecture  \cite{garnett20183dlanenet} with a ResNet34 \cite{ResNet} backbone. Our BEV projection covers 20.4m x 80m divided in the last decimated feature map to our tile grid $G_{W\times H}$ with $W=16, H=26$ such that each tile represents $1.28m \times 3m$  of road surface. We found that predicting the camera angle $\varphi_{cam}$ and height $h_{cam}$ gave negligible boost in performance compared to using the fixed mounting parameters on \textit{3D-lanes}, however, on \textit{synthetic-3D-lanes} we followed \cite{garnett20183dlanenet} methodology and trained the network to output $\varphi_{cam}$ and $h_{cam}$ as well.

The network is trained with batch size 16 using ADAM optimizer, with initial lr of 1e-5 for 80K iterations which is then reduced to 1e-6 for another $50K$ iterations. We set $\Delta_{pull}$ and $\Delta_{push}$ (Eqs. \ref{Eq:pull}, \ref{Eq:push}) to 0.1 and 3 respectively, and used a coarse 0.3 threshold on the output segment scores $\widetilde{c}_{ij}$ prior to the clustering. In the evaluation we set a coarse distance threshold for association of $1m$, and measured the AP as the average of $AP_{IOU\%}$ at IOU thresholds $0.1:0.1:0.9$.  

%% file: results.tex
\section{Results}\label{sec:results}

\begin{figure}[ht]
\vspace{-3mm}
\centering
\includegraphics[width=\textwidth]{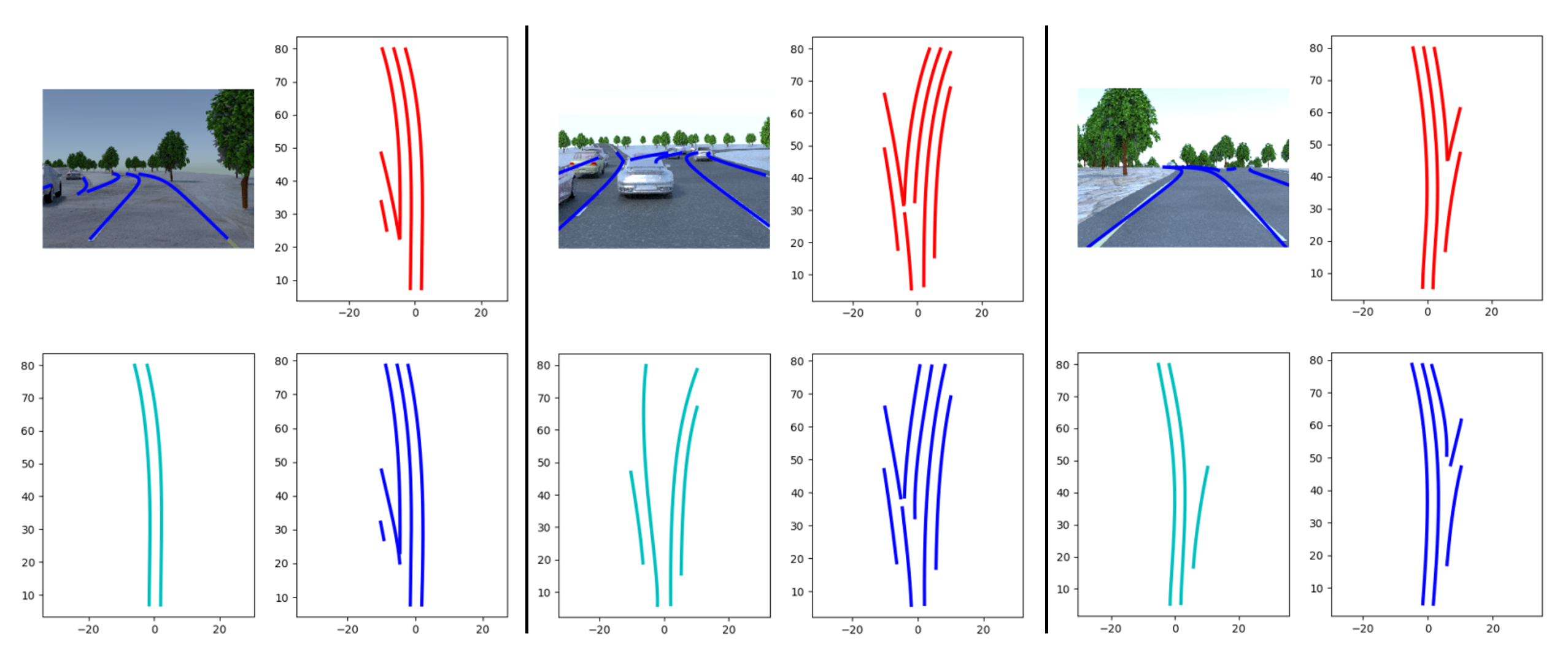}
\caption{Example results on \textit{synthetic-3D-lanes}. Our detected lanes \textcolor{blue}{(blue)}, the ground truth \textcolor{red}{(red)} and 3d-LaneNet lanes \textcolor{cyan}{(cyan)}. It is visible that our method is less constrained and detects all lanes in the scene.}
\label{fig:blender15}
\vspace{-5mm}
\end{figure}

\pbf{Synthetic-3D-lanes dataset }
We compared our method to 3D-LaneNet \cite{garnett20183dlanenet}. Results are presented in Table \ref{Table:blender}. It can be seen that our $AP$ and $AP_{50}$ are far superior to those of 3D-LaneNet\footnote{Results reported here differ than those in \cite{garnett20183dlanenet} since their evaluation disregards short lanes that start beyond 20m from the ego vehicle.} while showing comparable lateral error (for $IOU=0.5$ and $recall=0.75$). We believe the main reason is our semi-local representation that allows our method to support many different lane topologies such as short lanes, splits and merges that emerge only at a certain distance from the ego vehicle. This is evident in Fig. \ref{fig:blender15} showing examples where splits and short lanes are not detected by 3D-LaneNet but detected by our method. 


\begin{table}[!htbp] 
\vspace{-2mm}
\centering
\caption{comparison on \textit{synthetic-3D-lanes}}
\label{Table:blender}
\begin{tabular}{||c||cc||c|cc||}
\hline
Method & $AP$ &  $AP_{50}$ & recall &\multicolumn{2}{c||}{Lateral error (cm)} \\
\cline{5-6}
 & & & & 0-30m &  30-80m\\
\hline
3D-LaneNet \cite{garnett20183dlanenet}   &  0.74 & 0.79  &0.75 &\textbf{9.3}  & \textbf{23.9} \\
Ours   &  \textbf{0.9} & \textbf{0.95}  &0.75 & 9.7  & 26.7 \\
\hline
\end{tabular}
\vspace{-2mm}
\end{table}

\begin{figure}[ht]
\vspace{-2mm}
\centering
\includegraphics[width=\textwidth]{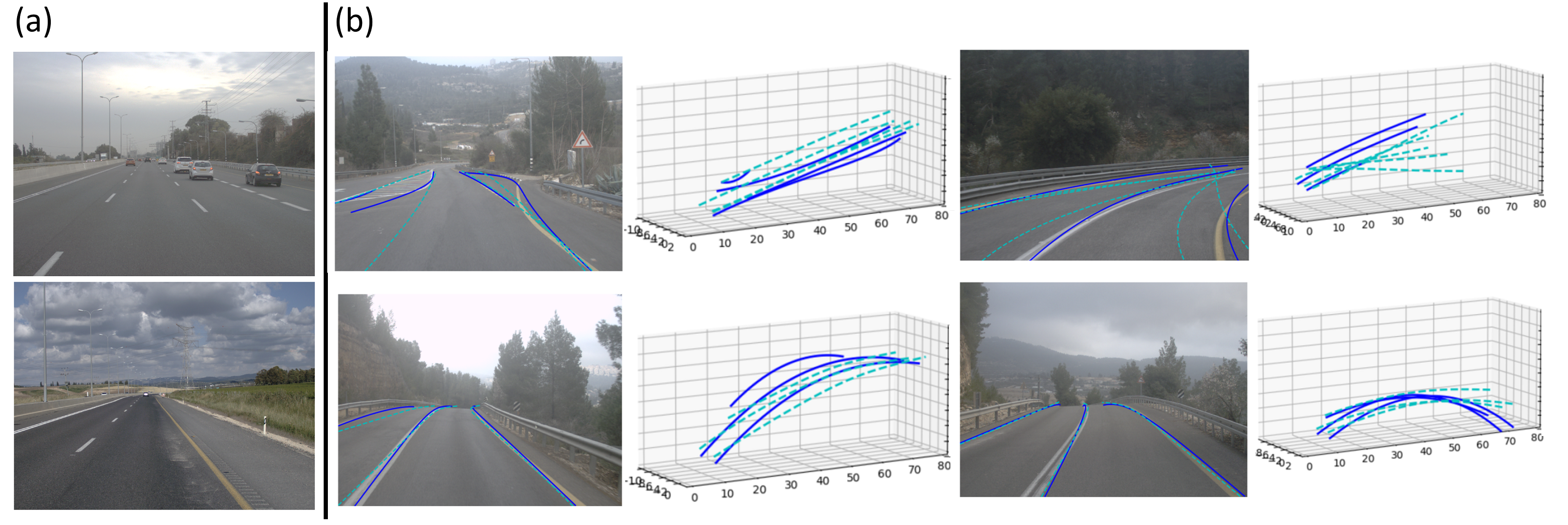}
\caption{Example results on \textit{3D-lanes}. (a) Depicts examples from the training set, and (b) examples from the test set. It is clear that the surface geometries and curvatures appear in the test set are different from the train set. Our detected lanes are shown in  \textcolor{blue}{blue} and 3D-LaneNet lanes in \textcolor{cyan}{cyan}.}
\label{fig:real_data}
\vspace{-5mm}
\end{figure}

\pbf{Generalizing to new topologies and geometries} We use the \textit{3D-lanes} dataset to demonstrate our methods ability to generalize to new scenes with complex curvatures and surface geometries. Results comparing our method to \cite{garnett20183dlanenet}, trained on the same train set, are summarized in Table \ref{Table:real_data}. 
It can be seen that our method achieves better results improving by 9 points over 3D-LaneNet in overall $AP$ as well as lowering the lateral error for the 3d lane points. This experiment is challenging compared to the \textit{synthetic-3D-lanes} experiment in which train and test sets have the same distribution. In the case of \textit{3D-lanes}, train and test sets have much different distribution with the test set exhibiting much more complex curvatures and surface geometries as also shown in the examples in Fig. \ref{fig:real_data}. 
We believe our ability to generalize to this test set demonstrates the advantage of using the proposed semi-local tiles representation.

\begin{table}[!htbp] 
\vspace{-2mm}
\centering
\caption{comparison on \textit{3D-lanes}}
\label{Table:real_data}
\begin{tabular}{||l||ccc||c|cc||}
\hline
Method  & $AP$ &  $AP_{50}$ & $AP_{90}$&  Recall & \multicolumn{2}{c||}{Lateral error (cm)}\\
\cline{6-7}
 & & & & & 0-30m &  30-80m\\
\hline

3D-LaneNet \cite{garnett20183dlanenet}   &  0.80 & 0.86 & 0.48   & 0.85 & 15.6  & 47.2 \\
Ours - w/o global  &  0.84 & 0.94 & 0.43 & 0.85 & 14.5  & 45.5 \\

Ours   &  \textbf{0.89} & \textbf{0.95} & \textbf{0.60}  & 0.85 &\textbf{14.1}  & \textbf{44.7} \\

\hline
Ours - w synthetic  & \textbf{0.9}  & \textbf{0.95} & 0.59 & 0.85 & \textbf{12.9}  & \textbf{36.3} \\
\hline
\end{tabular}
\vspace{-2mm}
\end{table}

To show the significance of our clustering approach using global feature embedding we compare it with a na\"ive clusterring alternative (Table \ref{Table:real_data} 'Ours - w/o global'). This alternative uses a simple greedy algorithm concatenating segments based on continuity and similarity heuristics. We find that in using na\"ive clusterring we loose 5 points in overall $AP$ and 17 in $AP_{90}$ suggesting that detected lanes with greedy clustering are much shorter. In addition, we see that with feature embedding we obtain lower lateral error. This may suggest that feature embedding learning also helps predicting more accurate segments. 

We also compare a model trained only on \textit{3D-lanes} with a one trained on both \textit{3D-lanes} and \textit{synthetic-3D-lanes} (Table \ref{Table:real_data} 'Ours -w synthetic'). We find that additional 3D training data of complex curvatures and geometries helps in the generalization despite it being synthetic and without using any domain adaptation techniques.


\begin{figure}[ht]
\vspace{-3mm}
\centering
\includegraphics[width=\textwidth]{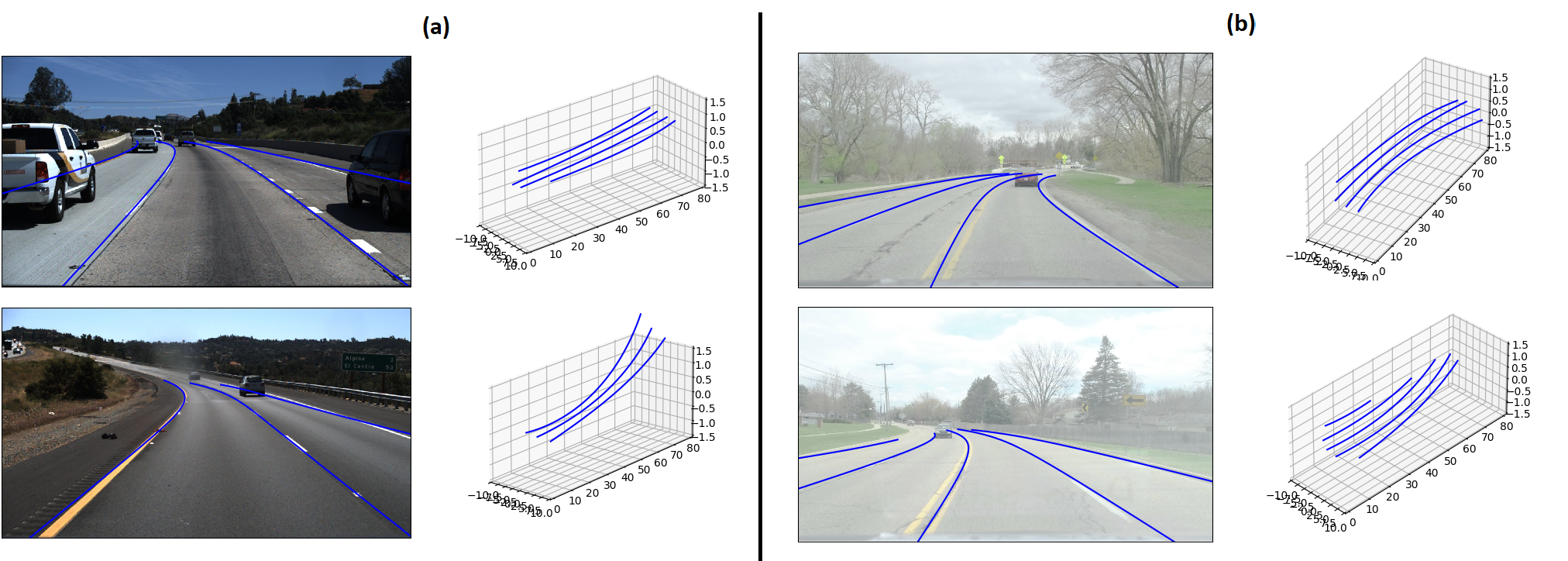}
\caption{generalization to new cameras: (a) depicts examples from $tuSimple$ test set (b) examples from our \textit{internal evaluation dataset}. Our method generalizes well to new cameras that weren't present in the training set without any need for adaptation. }
\label{fig:generalization}
\vspace{-6mm}
\end{figure}

\pbf{Generalization to new cameras} We now examine our methods generalization to new unseen cameras. To this end, we first use the 2D lanes tuSimple dataset \cite{tusimple_dataset} and show that our network, trained on a different task (3D lane detection rather than 2D), and on different data (\textit{synthetic-3D-lanes} and \textit{3D-lanes}), can generalize to a new task on unseen cameras. For inference on tuSimple we use fixed camera angle $\varphi_{cam}$ and height $h_{cam}$ from \cite{garnett20183dlanenet} to projected the feature maps to BEV.  The resulting tuSimple accuracy ($acc$) metric for this experiment, 
is 0.912. This result is surprisingly high given that our network is designed for detecting lanes in 3D, and more importantly, was not trained on a single example from the tuSimple dataset (See Fig. \ref{fig:generalization}a). Encouraged by this result, we next trained our network on the tuSimple dataset, lifting 2D lanes to 3D using a flat world assumption. When the flat world assumption is violated, the lifted 3D lanes no longer have the BEV properties of real 3D lanes, such as parallel lanes of constant distance between them, making this approach more challenging than solving the 2D detection problem directly. Once trained on tuSimple data the network reached an accuracy of 0.956 which is comparable to the 0.966 SOTA results of \cite{Hou_2019_ICCV}. We also conduct a qualitative evaluation on another unseen camera using an \textit{internal evaluation dataset} not used in training (See Fig. \ref{fig:generalization}b), Here too, we find good generalization to new cameras and scenarios.

\begin{figure}[!htbp]
\vspace{-2mm}
\centering
\includegraphics[width=\textwidth]{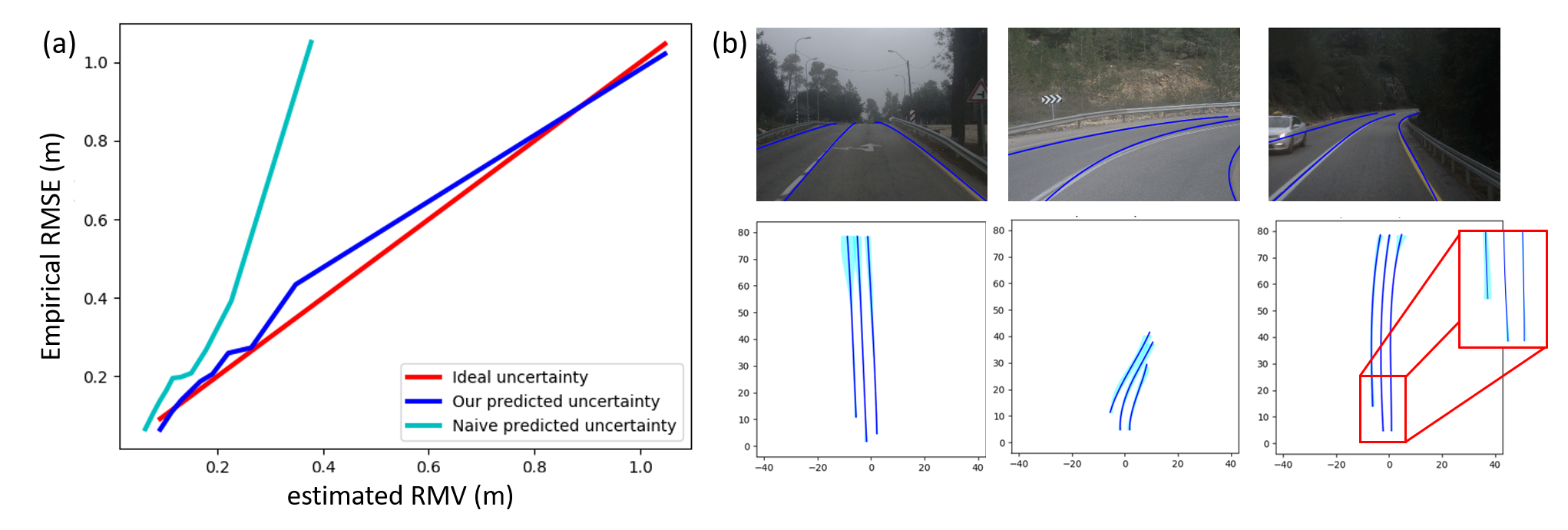}
\caption{uncertainty estimation results. (a) Shows comparison between our estimated uncertainty and a simple uncertainty baseline. (b) Examples of our predicted uncertainty. It is visible that the estimated uncertainty is larger on large curvatures (either lane curvature or surface curvature), and on occlusions (see bottom right inset) }
\label{fig:uncertainty_results}
\vspace{-5mm}
\end{figure}

\pbf{Uncertainty estimation} 
Our estimated uncertainty (variance) tries to quantify the noise in our detection model. That is, the empirical error between detected lane points and GT. Such noise is statistical by nature, thus can not be evaluate for a single sample (lane point). We do however, expect that the estimated uncertainty \textit{on average} would reflect the average empirical error. Therefore, in order to evaluate the estimated uncertainty we first divide it to bins, and compare the Root Mean estimated Variance (RMV) in each bin to the Root Mean Squared empirical Error (RMSE) of the samples (lane points) in that bin. Equality between the two measures indicates well calibrated uncertainty. This evaluation method is described by Levi et al. \cite{levi2020evaluating} which also propose a single figure of merit, the Expected Normalized Calibration Error (ENCE) which averages the error between the RMV and the RMSE in each bin, normalized by the bin’s RMV. 
Fig. \ref{fig:uncertainty_results}a shows the results for the above evaluation. In this analysis RMV is take as the maximal eigenvalue of our 3D covariance matrix which is reasonable since most of the error is in the lateral direction. We compare it to the empirical lateral RMSE taken between each detected point and the GT lane.
We can see that our estimated uncertainty is very close to the ideal uncertainty line achieving an ENCE of $11\%$. 

In the absence of a previous baseline, we compare ourselves to an uncertainty model similar to ours only it is supervised by errors computed in the tile level, i.e. on the same tiles we train our segment parameters (orange tiles in Fig. \ref{fig:tiles_MSE}a). As discussed in Sec. \ref{sec:method_uncertainty} we expect these errors to be bounded and generate a skewed distribution. Fig. \ref{fig:uncertainty_results}a) demonstrated that indeed this is the case. The maximal estimated RMV reaches $\sim40\%$ of the maximal empirical RMSE, and the resulting ENCE for this model is $60\%$. Fig. \ref{fig:uncertainty_results}b shows qualitative results of the estimated uncertainty. We can see that the uncertainty captures detection errors occurring at large curvatures (either lane curvature or complex surface geometries), occlusions, or large distance. Note that other approaches based on offline error modeling and rule based look up tables wouldn't necessarily capture all the cases in which the uncertainty should be high. This is in contrast to our data driven approach.

%% file: conclusions.tex
\section{Conclusions}
We presented a novel 3D lane detection with uncertainty estimation framework. The method uses a semi-local representation that captures topology-invariant lane segments that are then clustered together using a learned global embedding into full lane curves. The efficacy of our approach was showcased in extensive experiments achieving SOTA results while demonstrating its ability to detect globally complex lanes having different topologies and curvatures and generalize well to unseen complex surface geometries and new cameras. In this work we also implemented the first learning based uncertainty estimation for lane detection. We show the importance of properly quantifying the detection errors, achieving almost ideal uncertainty results with respect to the real error statistics.
Our work performs full 3D lane detection and uncertainty estimation thus closing the gap towards full lane detection requirements for autonomous driving.